\documentclass[preprint,12pt]{elsarticle}

\usepackage{amssymb}
\usepackage{amsmath}
\usepackage{amsmath,graphicx}
\usepackage{pifont,bbding}
\usepackage{graphicx}
\usepackage{amsmath}
\usepackage{amssymb}
\usepackage{booktabs}
\usepackage{adjustbox}
\usepackage{array}
\usepackage{amsmath,amsfonts}
\usepackage[table,xcdraw]{xcolor}
\usepackage{multirow}
\usepackage{hyperref}
\usepackage{float} 
\usepackage{caption}

\journal{Engineering Applications of Artificial Intelligence}

\begin{document}

\begin{frontmatter}



\title{NDLPNet: A Location-Aware Nighttime Deraining Network and a Real-World Benchmark Dataset}

%

\author[A]{Huichun Liu} 
\ead{Feecuin@outlook.com}

\author[A,B]{Xiaosong Li\corref{cor}}
\ead{lixiaosong@buaa.edu.cn}

\author[A]{Yang Liu}
\ead{by1917024@buaa.edu.cn}

\author[B]{Xiaoqi Cheng}
\ead{chexqi@163.com}

\author[A]{Haishu Tan}
\ead{tanhaishu@fosu.edu.cn}

\cortext[cor]{Corresponding author}
\affiliation[A]{organization={Guangdong-HongKong-Macao Joint Laboratory for Intelligent Micro-Nano Optoelectronic Technology, School of Physics and Optoelectronic Engineering, Foshan University},
            city={Foshan},
            postcode={528225},
            country={China}}
\affiliation[B]{organization={Guangdong Provincial Key Laboratory of Industrial Intelligent Inspection Technology, Foshan University},
            city={Foshan},
            postcode={528225}, 
            country={China}}
\begin{abstract}
Visual degradation caused by rain streak artifacts in low-light conditions significantly hampers the performance of nighttime surveillance and autonomous navigation. Existing image deraining techniques are primarily designed for daytime conditions and perform poorly under nighttime illumination due to the spatial heterogeneity of rain distribution and the impact of light-dependent stripe visibility. In this paper, we propose a novel \textbf{N}ighttime \textbf{D}eraining \textbf{L}ocation-enhanced \textbf{P}erceptual \textbf{Net}work(\textbf{NDLPNet}) that effectively captures the spatial positional information and density distribution of rain streaks in low-light environments. Specifically, we introduce a Position Perception Module (PPM) to capture and leverage spatial contextual information from input data, enhancing the model’ s capability to identify and recalibrate the importance of different feature channels. The proposed nighttime deraining network can effectively remove the rain streaks as well as preserve  the crucial background information. Furthermore, We construct a night scene rainy (NSR) dataset comprising 900 image pairs, all  based on real-world nighttime scenes, providing a new benchmark for nighttime deraining task research. Extensive qualitative and quantitative experimental evaluations on both existing datasets and the NSR dataset consistently demonstrate our method outperform the state-of-the-art (SOTA) methods in nighttime deraining tasks. The source code and dataset is available at https://github.com/Feecuin/NDLPNet.
\end{abstract}


\end{frontmatter}



\section{Introduction}
Image deraining is a fundamental yet challenging task for image restoration. The presence of rain significantly disrupts human activities various computer vision applications, particularly under nighttime conditions combined with inadequate lighting. These challenges are especially pronounced in tasks such as autonomous driving\cite{auto1,auto2}, object detection\cite{dete1,dete2}, segmentation\cite{seg1,seg2,seg3}, and photography. Rain streaks in a night scene are difficult to distinguish in an image due to the lack of light and the unpredictable location and number of light sources, resulting in clearly visible rain streaks close to the light source and blurred rain streaks away from the light source.This ambiguity can compromise the interpretation of scene features, affecting the decision-making processes of both humans and artificial intelligence, and potentially leading to severe security risks.

The emergence of rain removal method helps to alleviate the visual interference of rain in the image. In the early stage, people empirically used the characteristics of rainfall to deduce formulas, trying to separate the rain layer from the background layer. However, the robustness of this method based on manual derivation is often not ideal for complex rainy scenes, and its limitations are obvious. In recent years, the development of deep learning for rain removal methods is very rapid, and many methods based on deep learning have sprung up. Yang et al.\cite{binary} use binary rain masks to detect the location of rain, and then gradually remove rain through a recursive framework. Because the convolution mechanism of convolutional neural network determines that it has a small receptive field and can effectively extract local details, many rain removal methods using CNN architecture have appeared. However, convolutional neural network may pay too much attention to local details in images, resulting in performance degradation. The deep learning rain removal method based on Transformer can overcome the neglect of global information due to its excellent global modeling ability. Although these methods have achieved good results,the development of rain removal methods has been instrumental in alleviating the visual interference caused by rain. 

The biggest difference between the night scene and the day scene of rain is that there is no unified light source illumination, which leads to different appearances and discrimination difficulties of rain streaks in different spatial positions on the same image containing rain due to the differences in light source illumination. The rain streaks near the light source are clearly visible, highly identifiable and densely distributed, while the rain streaks far away from the light source are vice versa.
Fan et al.\cite{nightmethod} believe that it is more important to identify the location information of rain streaks in the night scene than in the day scene. But they only simply extract the information of rain streaks and do not emphasize the importance of the location of rain streaks in space and the distribution of rain streaks density in the image, which may lead to the loss of some details or blur in the image processed by the network, and the residue of rain streaks.

There are two predominant issues: first, the majority of existing deep learning-based rain removal methods\cite{in-method1,in-method2,in-method3,in-method4,in-method5} are mainly designed  for daytime scenes, and their application is limited in nighttime conditions, thus leaving a big gap in this field. Second, commonly used rain datasets\cite{dataset1,dataset2,dataset3} scarcely include night scenes. This hinders the direct application of existing methods to nighttime deraining. Although some recent datasets\cite{night1,night2} have begun to incorporate night scenes, the number and diversity of night scenes compared to daytime scenes are still insufficient. 
These two problems lead to the popular rain removal method not having as good performance when dealing with nighttime scenes as it does with daytime scenes. 

In order to solve the two problems that arise above, in this paper, we propose a location-enhanced perceptual rain removal method for rain streaks in nighttime scenes, named as NDLPNet. Firstly, the location of rain streaks in the rain image was initially obtained by implicit learning of the cyclic residual model using\cite{nightmethod} rainfall location prior. The obtained rain grain prior map was input into the position perception module to generate spatial location embedding, which helped the model to better capture and use the spatial context information of the input data, and more efficiently use the channel information in the feature map. The model's ability to identify and adjust the importance of each channel in the feature map was further enhanced, and attention to irrelevant information was reduced. As a result, the model can better retain the original image content details, thereby improving the overall performance. In addition, we construct a semi-realistic nighttime rain dataset (NSR) covering 900 image pairs, Compared to previous datasets, NSR dataset makes the nighttime scene data more diverse and closer to the real scene at night, and  further makes up for the lack of nighttime rain dataset.

The main contributions can be summarized as follows:
\begin{itemize}
    \item This paper introduces a novel rain-removal method for nighttime scenes, designed to generate high-quality rain-free images from nighttime rainy conditions while preserving fine details and textures.
\end{itemize}
\begin{itemize}
    \item A position perception module(PPM) is proposed to further capture the spatial location information and density distribution of rain streaks in the image, which helps the model to efficiently remove rain while better preserving the original image content details.
\end{itemize}
\begin{itemize}
    \item A semi-realistic nighttime rain-containing dataset is constructed, which is a synthetic dataset of rain in real scenes at night, and is specially used for rain removal research at night, further making up for the shortcomings of rain datasets at night.
\end{itemize}
\begin{itemize}
    \item Extensive experimental results on existing datasets demonstrate the superiority of our method, and comprehensive ablation experiments also verify the effectiveness of each module of our work.
\end{itemize}

\section{Related Work}
\noindent\textbf{Single Image Deraining.} Image rain removal has always been a challenging and important task in computer vision and image processing. Image rain removal methods can be classified into traditional and learned categories. Early traditional image rain removal methods focus on the use of carefully designed priors to provide additional constraints and thus separate rain components, such as Gaussian mixture models\cite{gaosi}, sparse representation learning\cite{xishu1,xishu2}, and oriented gradient priors\cite{fangxiang}. However, such manually derived priors are derived relying on existing empirical knowledge, fail to model the inherent properties of clear images, often lack good generalization ability, and are usually difficult to adapt to diverse scenarios. 

With the rapid development of deep learning, various deep learning networks have been explored for image restoration\cite{EAAI-5,EAAI-6,ACCV,EAAI-4,EAAI-7}, and learning-based rain removal methods\cite{related-method1,related-method2,related-method4,top1,top2,top3,top4} have gradually become dominant in the field of rain removal. Fu et al.\cite{in-method5} proposed the seminal work deep residual network for image rain removal. In order to better represent the distribution of rainfall on the image, some studies start to solve the method from the direction, depth and density of rainfall, and optimize the network structure through some recursive strategies. Li et al.\cite{RESCAN} proposed a novel approach that integrates Squeeze-and-Excitation (SE) blocks to adaptively assign distinct \(alpha\) values to different rain streak layers based on their intensity and transparency levels. Xiao et al.\cite{IDT} proposed a converter system containing complementary window-based converters and spatial converters, thus enabling improved capture of both short-term and long-term dependencies in rainy day scenes. Chen et al.\cite{DRSformer} proposed a learnable top-k operator that selects the attention scores most relevant to the key. Additionally, they introduced a hybrid-scale feedforward network to perform multi-scale modeling of latent images. Chen et al.\cite{SR} proposed a sparse sampling converter based on uncertainty-driven ranking, which can adaptively sample relevant image degradation information to model the underlying degradation relationship. 
Tao et al.\cite{EAAI-3} proposed a dual-stage network based on clean image detail reconstruction, which aggregates both high-level and low-level rain features along with contextual information. By employing loss functions that emphasize image edges and regional accuracy, the method achieves structurally precise rain removal effects.
Chen et al.\cite{EAAI-2} proposed an network that achieves parameter sharing in deraining tasks through feature-wise rain streak disentanglement. The framework incorporates a bidirectionally invertible channel interaction mechanism to enhance cross-channel information flow, thereby optimizing the separation of rain artifacts from underlying image content.
Wang et al.\cite{EAAI-1} proposed a domain adaptive deraining network to address the domain gap between synthetic and real-world rain patterns. The framework generates photorealistic rain layers through feature affine transformation guided by real rain streak information, while employing a joint training strategy combining contrastive learning and collaborative learning to achieve high-quality rain removal.
Song et al.\cite{ESDNet} applied SNN to the task of image de-raining, By learning the fluctuation of rain streaks to effectively detect and analyze the characteristics of rain streaks, the gradient surrogate strategy is introduced to directly train the model, which can better guide the rain removal process.

However, these methods are basically designed for daytime rainfall, and do not take into account some characteristics such as complex and variable location and appearance of rain streaks in nighttime scenes, and uneven density distribution. Therefore, the performance of rain removal in nighttime scenes will not be as excellent as that in daytime scenes. Fan et al.\cite{nightmethod} proposed a prior to extract rain streaks information for night scenes, but did not emphasize the importance of the location of rain streaks in space and the distribution of rain streaks density in images. In this paper, an enhanced perception method of rain streaks location is proposed for nighttime scenes, and spatial location coding is used to further capture the information of rain streaks to achieve accurate rain removal. In this work, we compare with some recent rain removal methods in nighttime scenarios.\\
\textbf{Vision Transformer.} Due to the great success of Transformers in the field of natural language processing and advanced tasks, more and more image restoration tasks are being designed using Transformers\cite{in-method2,in-method3,vit1,vit2,vit3}. In the field of image rain removal,  Zamira et al.\cite{related-method4} proposed MPRNet, which makes full use of encoder-decoder architecture and multi-stage strategy. They also proposed Restormer\cite{vit2}, which uses a Transformer model to efficiently process high-resolution images. Chen et al.\cite{DRSformer} proposed the learnable top-k selection operator to adaptively maintain the most useful self-attention value for better feature aggregation. 
Liang et al.\cite{in-method2} proposed the Deraining Recursive Transformer (DRT), which employs a recursive local-window based self-attention architecture to significantly reduce computational resource consumption while maintaining the performance advantages of Transformers, effectively addressing the challenges of parameter-heavy existing Transformer models in dense prediction tasks like deraining and their difficulty in deployment on resource-constrained devices.
Although the above Transformer based rain removal methods have good performance and robustness, the general Transformer can only process one-dimensional text and two-dimensional image data, and cannot capture the complex spatial information of three-dimensional images. It is very difficult to determine the location of rain streaks in the night scene due to the lack of depth information. Therefore, the above method cannot remove rain streaks in the night scene as well as in the day.\\
\textbf{Dataset.} With the rise of image rain removal tasks, more and more rain removal model methods based on deep learning have emerged. Most data-driven rain removal models require pairs of rainy images and clean, rain-free ground truth for training. Since real rain image pairs are difficult to collect, the early rain datasets\cite{12000,14000, outdoor-rain,rain100,rain800} is usually a synthetic dataset. Although the synthetic images of these datasets contain some physical features of real rainfall, the differences between the synthetic and real domains result in a large gap between the synthetic and real data. The later dataset\cite{GT-rain} addresses some of the inherent shortcomings of the synthetic dataset, but the above dataset is based on the daytime de-rainfall domain, and the de-rainfall dataset in nighttime scenarios has yet to be addressed.

In recent years, there have been some rainy datasets containing nighttime scenes\cite{night1,night2}, which make up for the lack of nighttime scenes in rainy datasets, but there is still a large gap between the quantity and quality of daytime datasets. Additionally, the existing nighttime rain datasets are almost a virtual scenes, and none of the images contain the noise that often accompanies night scenes, which also has a big gap with the real scene. Therefore, this paper proposes a semi-realistic nighttime rain image dataset, covering a variety of nighttime real scenes, to make up for the shortcomings of rainy datasets in nighttime scenes.

\section{Proposed Method}
\begin{figure}[h]
    \centering
    \includegraphics[width=1\linewidth]{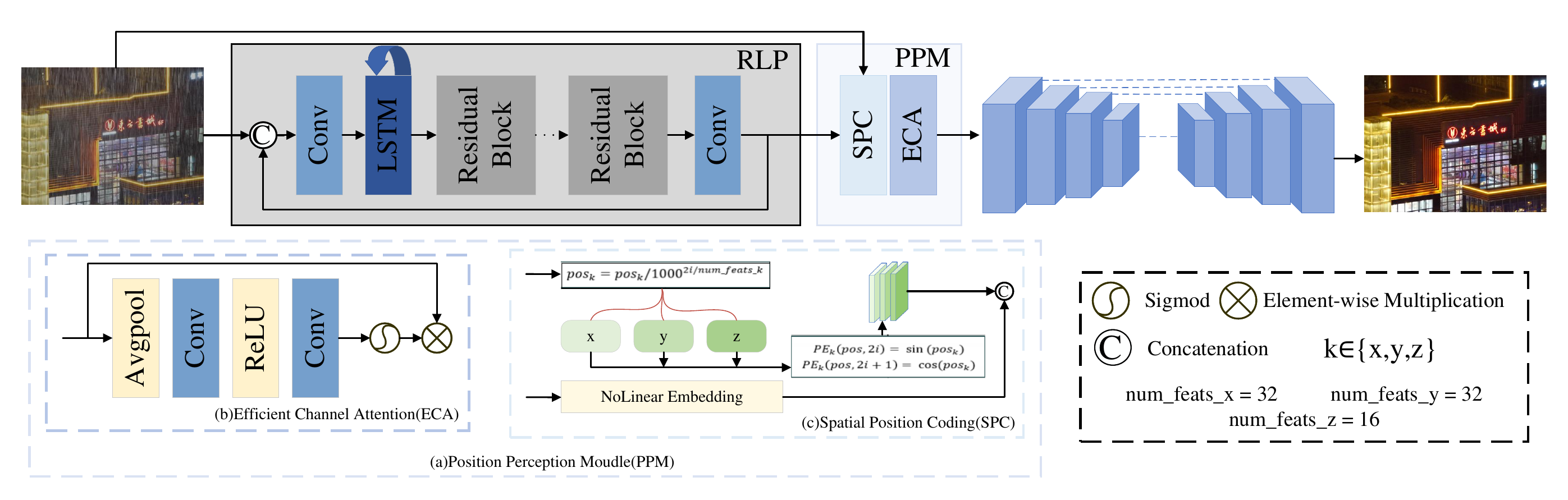}
    \vspace{-1em}
    \caption{Overview of our NDLPNet framework. Among them, the position perception module(PPM) goes through four layers of Restormerblock after coming out, and the input dimension and maximum depth are adjusted to 128 and 1024, respectively.}
    \label{fig1}
\end{figure}

In this section, we describe the architecture and framework of our proposed method, followed by detailed discussions on the individual modules within the architecture. The framework diagram is illustrated in Fig.\ref{fig1}.
\subsection{Overall pipline}
As depicted in Fig.\ref{fig1}, our network architecture accepts rain images as input and processes them to produce high-quality restored images as output. Given a rainfall image \( I_r \in \mathbb{R}^{H \times W \times C} \), where \(H \times W\) represents the image resolution and \(C\) is the number of channels, the image undergoes several transformations within the network. Initially, the image is processed through a Rain Location Prior (RLP) and then embedded with overlapping patches using a \(3 \times 3\) convolution. Subsequently, the Position Perception Module (PPM) is employed to further extract and emphasize the rain streaks' location, enhancing significant features while suppressing irrelevant information. The processed data is then forwarded through four layers of Restormer blocks for advanced learning. Pixel unshuffle and pixel shuffle operations are applied for feature downsampling and upsampling, respectively, ensuring the reconstruction of a high-quality, crisp output.
\subsection{Rain Location Prior}
We use the Rain Location Prior(RLP)\cite{nightmethod} as the prior of the network, which is used to extract the preliminary location information of rain streaks in the rain image.
The Rain Location Prior (RLP)\cite{nightmethod} incorporates residual blocks within a recurrent residual learning framework. These blocks effectively model rain streaks as residual components and progressively capture rain-specific high-frequency features through a recursive design. Starting with a rainy input and an initial prior map, the module iteratively extracts deep features and updates the prior map cyclically as shown below:
\begin{equation}
\bar{R}_k= F_{RLP}  (I \odot \bar{R}_{k-1}; \theta),k = 1,...,N,
\label{eq1}
\end{equation}

where \(I\) denotes the input, \(\bar{R}_k\) is the output of RLP at stage \(k\), and \(\bar{R}_0\) is the initial mapping set to 0.5. \(F_{RLP}(o; \theta)\) refers to the RLP module, \(\theta\) represents its parameters, \(\odot\) indicates a cascade of channels, and \(N\) is the total number of recursive stages, set to 6 in our method.
\subsection{Position Perception Moudle}
\begin{figure}[H]
    \centering
    \includegraphics[width=1\linewidth]{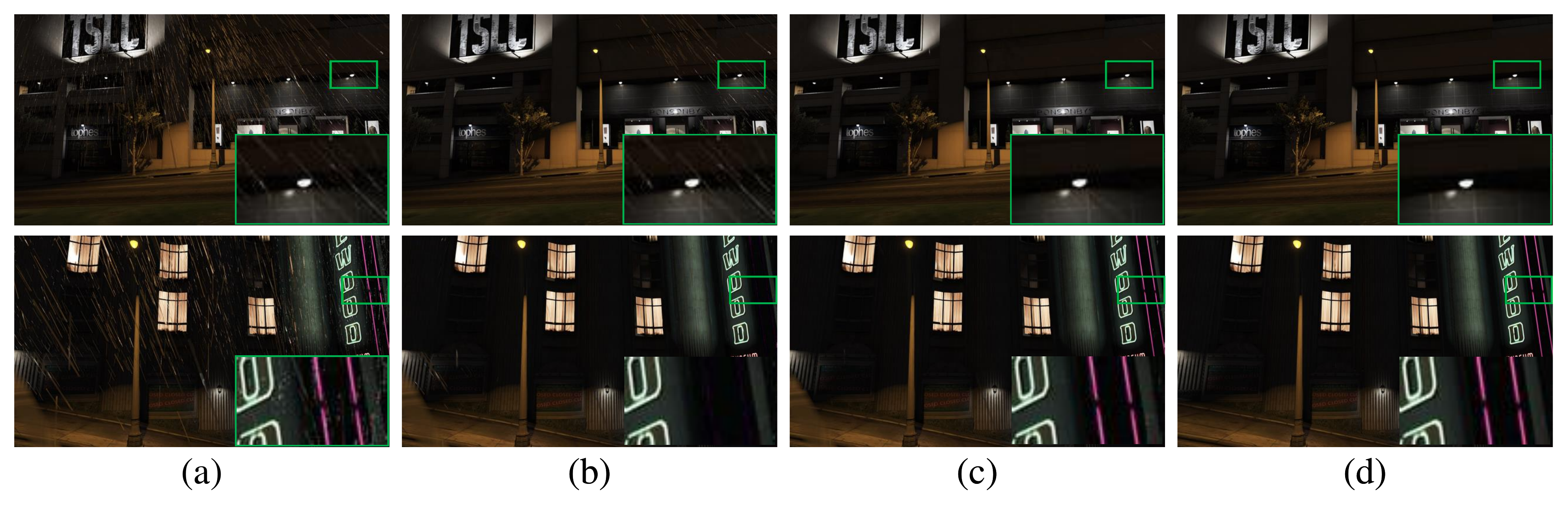}
    \vspace{-1em}
    \caption{(a) is the image with rain. (b) is the image processed only with RLP prior. (c) is the image processed after adding PPM module. (d) is the Groundtruth corresponding to the image with rain. (magnification can provide a clearer view)}
    \label{figk}
\end{figure}

As shown in (b) in Fig.\ref{figk}, although the locations of rain streaks can be initially obtained to help the model remove rain, the RLP prior does not focus on identifying or retaining important features or details of rain streaks. The lack of emphasis on the dense or sparse areas of rain streaks leads to insufficient rain streak feature extraction. The model pays attention to the rain streak information and also pays attention to the other irrelevant feature information, which leads to insufficient detail retention of the recovered image or the incomplete removal of rain streaks.

In order to solve this problem, we propose a position perception module to further capture the spatial location information and distribution of rain streaks in the rain image. At the same time, we enhance the ability of the model to identify and adjust the importance of each channel of the feature map, reduce attention to irrelevant information, and help the model better retain the original image details. The position perception module consists of spatial location encoding and efficient channel attention. As shown in Fig.\ref{figk}(c), the details of the original image are well preserved after the position perception module is added, and the rain is removed accurately. These two modules are described in detail below.
\subsubsection{Spatial Position Coding.} 
The location embedding method typically used by Transformer models is based on the logical sequence numbering of positions, rather than directly reflecting the physical spatial arrangement. This approach is highly effective for processing one-dimensional textual data, as it captures the relative order of words within a sentence. However, this method of embedding based on logical order encounters limitations when applied to binary or three-dimensional data, such as images. 
Traditional Transformers are unable to adequately represent complex spatial information, such as two-dimensional or three-dimensional structures, and might overlook crucial details like the actual distance between pixels. General image spatial position coding\cite{SPC1,SPC2,SPC3} is two-dimensional, which cannot capture the intensity of rain in different regions of the image. When processing images with rain at night, the network cannot focus on a certain region according to the intensity of rain in the image, which may cause the network to not only focus on the rain grain. At the same time, the network will not be able to self-adjust the concentration according to the density of rain streaks in a certain area, resulting in problems such as blurred details, distortion, or residual rain streaks in some areas.

Therefore, we propose a more refined and adaptable location encoding module to assist the Transformer in processing nighttime rain images, so that the model can understand and use the location information and distribution of the input image in multiple spatial directions, which enhances the model's ability to understand complex spatial layouts. The encoding formula is as follows:
\begin{equation}
{PE}_k(pos,2i)=sin({posv}_k),
\label{eq2}
\end{equation}
\begin{equation}
{PE}_k (pos,2i+1)= cos(posv_k),
\label{eq3}
\end{equation}
where \({PE}_k\) represents the encoded information in dimension \(k\), with \(k \in \{x, y, z\}\), and variable \(i\) denotes the position in the tag vector, which represents the index of the tag vector.
\begin{equation}
{posv}_k={pos}_k/1000^{(2i/(num\_feats\_k))},
\label{eq4}
\end{equation}
let denote the value of position encoding in dimension \(k\) and denote the number of position encoding features in the direction of dimension \(k\), where we set the number of features for \(x\), \(y\), \(z\) to 32, 32 and 16 respectively. The encoded information calculated on the three dimensions is concatenated together to form a complete position code. Finally, it is concatenated with the source image linearly embedded by a 3x3 convolution to enhance the spatial information of the feature map, so that the model can enhance the spatial perception ability and improve the performance of the model.

The module not only considers the two-dimensional space coordinates of \(x\) and \(y\), but also adds the rain grain density information according to the RLP prior. The rain grain density information is crucial to adjust the concentration of a certain region and emphasize the rain grain characteristics of that region, which can prevent the network from paying too much attention to irrelevant information and reduce the impact of redundant information. This encoding method enables the model to accurately capture the spatial relationship and distribution between pixels, which is crucial for the image rain removal task because the distribution of raindrops or rain marks in the image is often closely related to the spatial location. The fine position encoding helps the model learn the distribution pattern, size and shape of raindrops or rainmarks in the image and their interaction with the surrounding environment, so as to remove raindrop effects more accurately.
\subsubsection{Efficient Channel Attention.} 
Following the position encoding process, the resultant feature map is fed into the efficient attention module\cite{woo2018cbam}, which minimizes the impact of superfluous information while optimizing feature extraction. This module aids in filtering out irrelevant data and diminishes the model's focus on such distractions. The positional data provided by the encoding process allows the channel attention module to adjust channel weights based on the specific encoding of each pixel's position during the computation of attention weights. Consequently, even within a single channel, pixels at different locations may receive varied attention weights, reflecting their relative significance in a specific spatial context. The formula for efficient channel attention is presented below, where \(x\) represents the input feature map, \(y\) the output feature map, and \(A\) the channel attention weight:
\begin{equation}
A = \sigma(\text{Conv}(\text{ReLU}(\text{Conv}(\text{AvgPool}(x))))),
\label{eq5}
\end{equation}
\begin{equation}
y = X \odot A,
\label{eq6}
\end{equation}
where \(\text{Conv}\) denotes the convolution operation, \(\sigma\) represents the sigmoid activation function, and \(\odot\) denotes the Hadamard product. 
Efficient channel attention facilitates a more nuanced consideration of the interactions between channel features and the location information of rain streaks. This enhancement helps the model to concentrate on inter-channel relationships while fully acknowledging the influence of spatial factors on feature importance, thereby improving the representation of rain streak information and enhancing the model's processing and retention capabilities.
\subsection{NSR Dataset.}
\begin{figure}[H]
    \centering
    \includegraphics[width=1\linewidth]{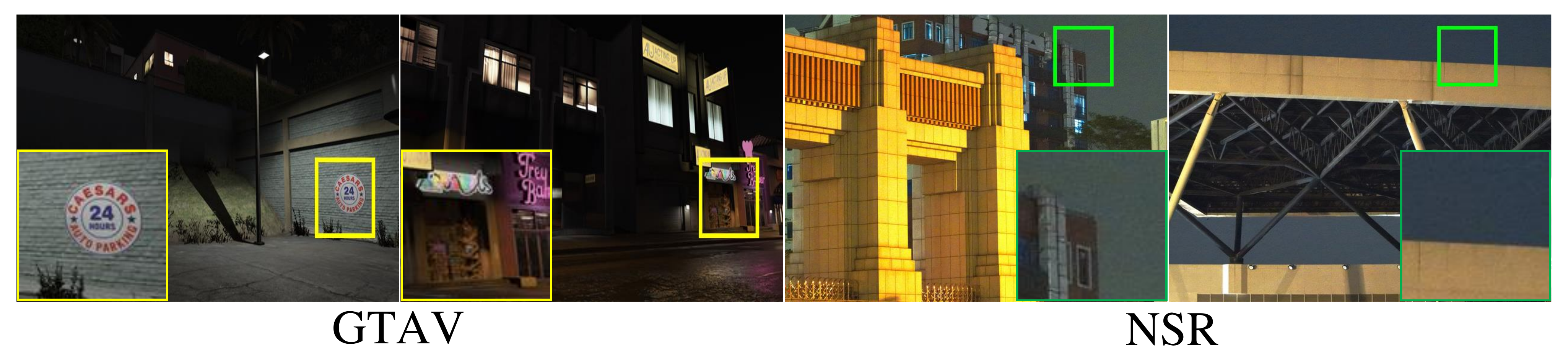}
    \vspace{-1em}
    \caption{Difference between NSR dataset and GTAV dataset regarding noise.}
    \label{fig21}
\end{figure}

The size as well as the richness of the dataset has a significant impact on the performance of the method. Most of the existing datasets with rain are based on daytime scenes, and the datasets about nighttime scenes are inferior to the datasets of daytime scenes in both quality and quantity. Moreover, the existing datasets with rain in nighttime scenes are all synthetic datasets, which ignore the gap between virtual scenes and real scenes, as shown in Fig.\ref{fig21}, real nighttime scenes are often accompanied by noise, which is not present in previous synthetic datasets, which hinders the development of nighttime rain removal methods. To this end, we construct a nighttime rain dataset based on real scenes to make up for the shortcomings of the datasets in nighttime scenes. Different from the previous fully synthetic nighttime rain dataset, our constructed dataset is based on real scenes shot manually, which narrows the gap with real rain data compared with the fully synthetic dataset.

\begin{figure}[h]
    \centering
    \includegraphics[width=1\linewidth]{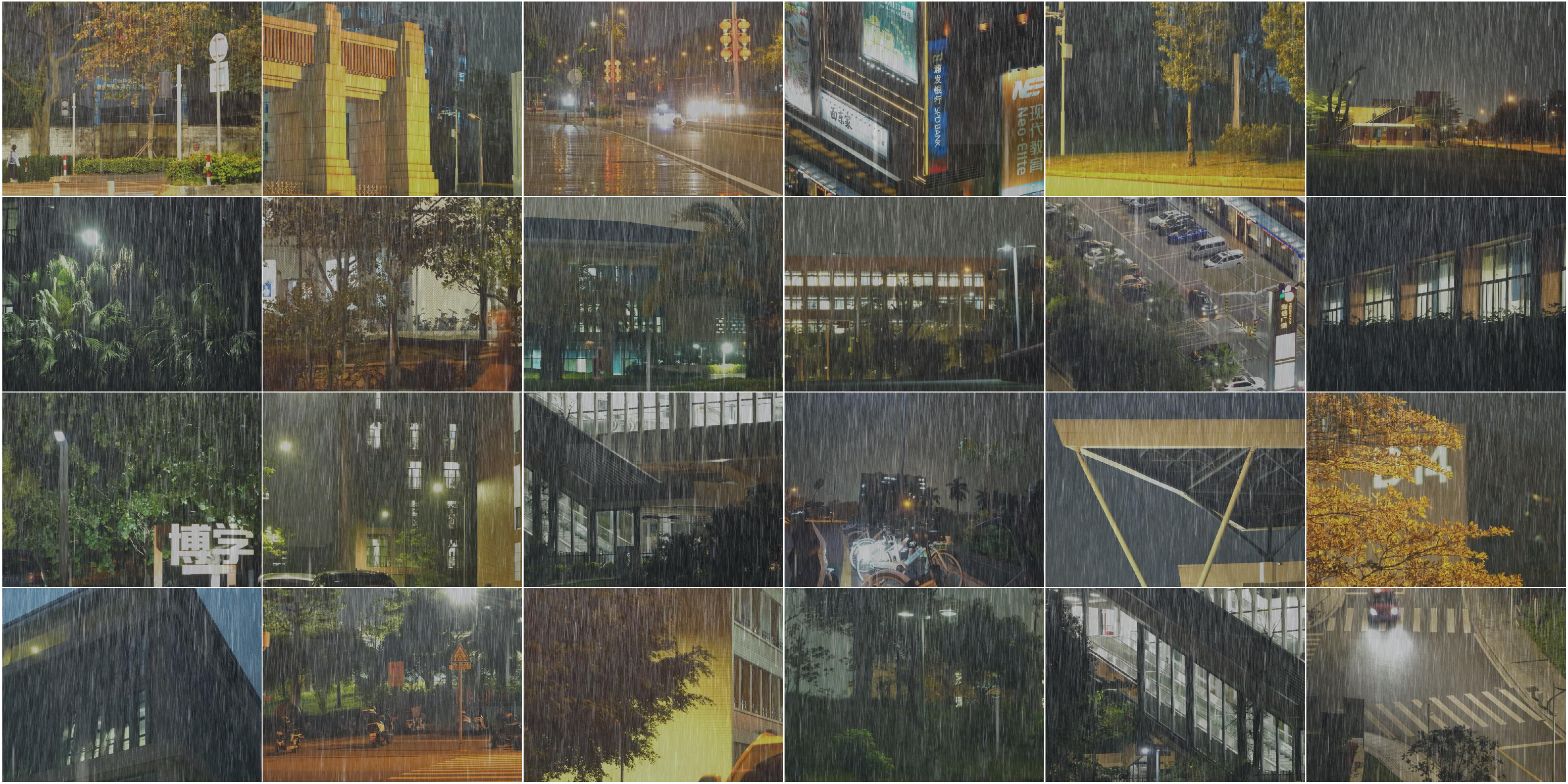}
    \caption{Part of our NSR dataset for night scene derain task.}
    \label{fig20}
\end{figure}

Taken by DJI M30T unmanned aerial vehicles, the dataset consists of 979 image pairs covering a variety of scenes, including campus, parking lot, park, urban road, stadium, subway station, etc., of which 900 image pairs are used for the training set and 79 image pairs are used for the test set. We used Photoshop software to generate masks based on the method\cite{dataset} to introduce rain grain details in the source images.
Fig.\ref{fig20} shows a partial presentation of the NSR dataset

In this paper, we utilize four layers of the Restormer block to reconstruct clear images. Each Restormer block comprises a Multidimensional Convolutional Head Transposed Attention (MDTA) module and a Gated Dconv Feedforward Network (GDFN). The MDTA module aggregates both local and non-local pixels and models the global context implicitly by applying self-attention within the channel dimension. This ensures efficient processing of high-resolution images. Meanwhile, the GDFN employs a dual-gating mechanism to suppress less informative features, allowing only pertinent information to advance through the network hierarchy, which is advantageous for focusing on relevant rain pattern features and restoring clean images.

\section{Experiment}
In this section, we introduce the setup of our experiments and then proceed to both quantitatively and qualitatively evaluate our method on synthetic datasets for night scenes, demonstrating its superior performance in rain removal compared to existing methods. Additionally, we assess the synthetic dataset under daytime conditions to verify the generalization capabilities of our model. We also perform ablation studies to validate the effectiveness of the proposed modules.
\subsection{Experimental Setups}
The method uses the pytorch architecture and is trained on Nvidia GeForce RTX 3090 GPUs, and all experiments were performed on a single Nvidia GeForce RTX 3090 GPU. The network architecture scales from level 1 to level 4 with the number of Transformer blocks set to [4, 6, 6, 8]. The initial dimension is configured at 128, expanding up to a maximum of 1024. The model was trained over 200K iterations using the AdamW optimizer (\({\beta}_1\)=0.9, \({\beta}_2\)=0.999, weight decay 1e-4) with L1 loss. To further assess generalization in daytime scenarios, training iterations were extended to 300K. The initial learning rate of 3e-4 was progressively decreased to 1e-6 through cosine annealing. Training patches were sized at 128x128, with a batch size of 2.\\
\textbf{Dataset.}
Experiments were conducted on the synthetic datasets Rain200L, Rain200H\cite{dataset3}, GTAV-NightRain\cite{night2} and our newly constructed NSR dataset. The GTAV-NightRain dataset comprises 10 rainy images and 1 corresponding clean image per scene, with distinct training and test sets featuring non-overlapping scenes. We utilized 2000 pairs for training and 100 pairs for testing from this dataset. The Rain200L and Rain200H datasets, designed for daytime scenarios, each contain 1800 training pairs and 200 testing pairs of synthetic rain images. The NSR dataset consists of 900 training pairs and 79 testing pairs of synthetic rain images from real scenes.\\
\textbf{Compared Methods and Metrics.}
To demonstrate the effectiveness of our approach, we conducted extensive experiments to compare various methods available for image rain removal. Firstly, we compare with PReNet\cite{related-method1}, Uformer\cite{vit1}, ESDNet\cite{ESDNet}, RLP\cite{nightmethod}, DiG-CoM\cite{fangxiang} and UNet\cite{UNet} on the synthetic nighttime rain dataset GTAV-NightRain and semi-real nighttime rain dataset NSR. We leverage the publicly available code for model retraining, and all methods are trained on synthetic datasets and our NSR dataset using the source code provided by the authors, following the default settings of the original authors. Model performance is then evaluated on the test dataset to ensure a fair comparison. 

To demonstrate the model's ability to generalize well in daytime scenarios, and to highlight the diversity of methods compared in our experiments, Other current SOTA methods for daytime rain removal (SPD-Net\cite{spd}, Dual-GCN\cite{dualgcn}, Restormer\cite{vit2}, IDT\cite{IDT}, DRSformer\cite{DRSformer}, ESDNet\cite{ESDNet}) are visually compared on the datasets Rain200L and Rain200H of daytime scenes. For the quantitative assessment of image quality, we use the commonly used PSNR and SSIM as the evaluation indicators of the above methods to quantify the rain removal performance of different methods.
\begin{figure}[h]
    \centering
    \includegraphics[width=1\linewidth]{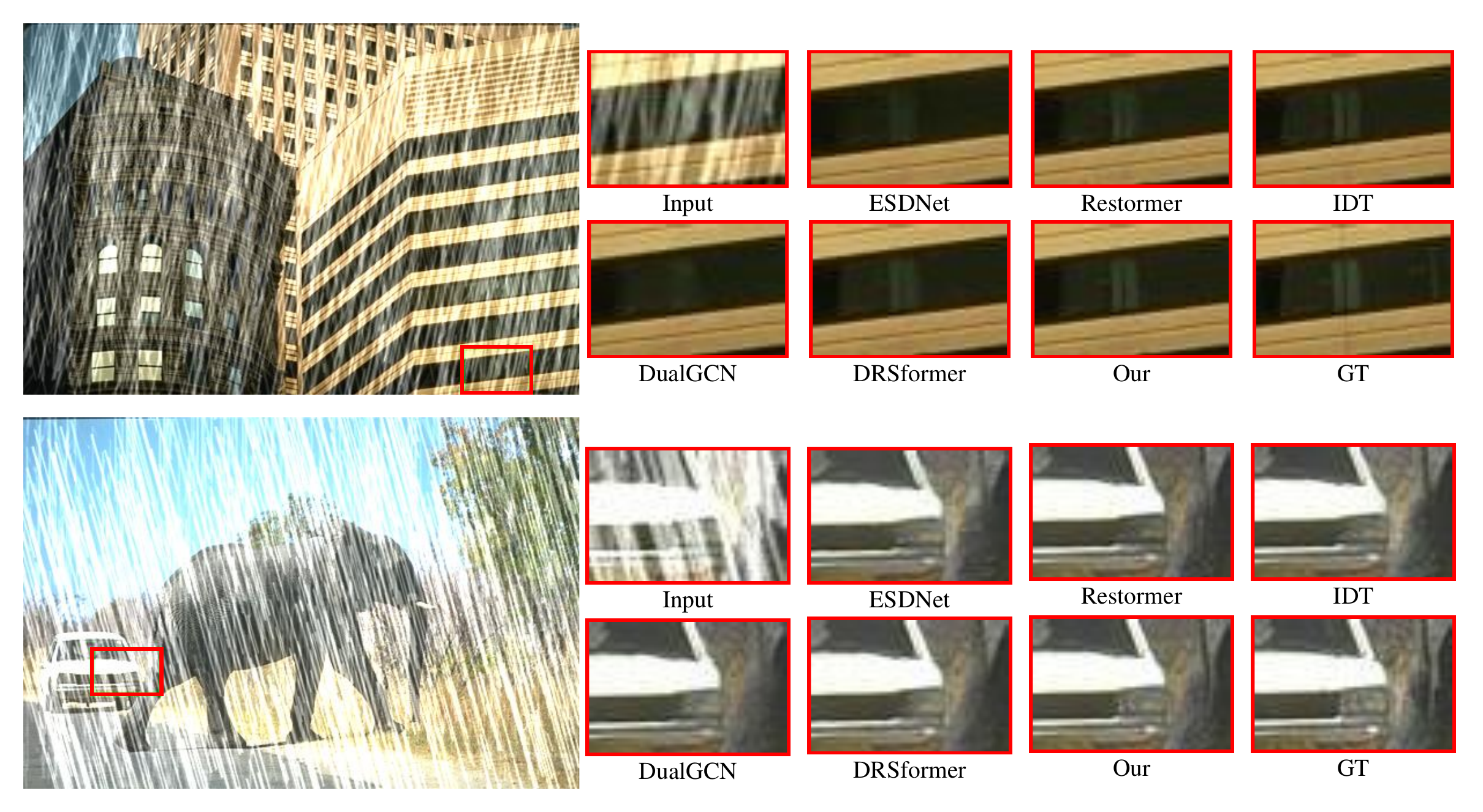}
    \caption{ Visual comparison of different deraining methods on Rain200H dataset (magnification can provide a clearer view).}
    \label{fig5}
\end{figure}

\begin{table*}
\centering
\caption{Quantitative comparison of various deraining methods trained on the GTAV-Nightrain and NSR dataset, \textbf{Bold} is the best, \textcolor{red}{red} is the second.}
\begin{adjustbox}{width=\textwidth}
\begin{tabular}{c c c c c c c c c c}
\toprule
\multirow{2}{*}{Dataset} & \multicolumn{1}{c}{Methods} & \multicolumn{1}{c}{Input} & \multicolumn{1}{c}{UNet\cite{UNet}} & \multicolumn{1}{c}{PReNet\cite{related-method1}}  & \multicolumn{1}{c}{DiG-CoM\cite{fangxiang}}   & \multicolumn{1}{c}{Uformer\cite{vit1}}& \multicolumn{1}{c}{RLP\cite{nightmethod}} &\multicolumn{1}{c}{ESDNet\cite{ESDNet}}  & \multicolumn{1}{c}{Our} \\
 & \text{Venue\&Year}  & - & \textit{(CVPR’2018)} & \textit{(CVPR'2019)} & \textit{(ICME'2020)} &\textit{(CVPR’2022)} & \textit{(ICCV'2023)} &\textit{(IJCAI'2024)} & - \\
\midrule
\multirow{2}{*}{GTAV-Nightrain} &  PSNR $\uparrow$        &29.15            & 35.45  & 34.77 & 30.55     & 35.59 &\textcolor{red}{36.19} & 36.85 &\textbf{38.68} \\ 
\multirow{2}{*}{}& SSIM $\uparrow$          &0.8202           & 0.9585   & 0.9484  & 0.8855     & 0.9639 &\textbf{0.9671} &0.9625   &\textcolor{red}{0.9661} \\ 
\midrule
\multirow{2}{*}{NSR} & PSNR $\uparrow$       &19.38             & 29.87   & \textcolor{red}{34.39} & 20.57     & 30.47 &31.35 &34.30 &\textbf{35.95}\\ 
\multirow{2}{*}{} &SSIM  $\uparrow$         &0.6247          & 0.9207   & \textcolor{red}{0.9478} & 0.7865     & 0.9270 &0.9316 &\textcolor{red}{0.9478} & \textbf{0.9524}\\
\bottomrule
\end{tabular}
\label{table1}
\end{adjustbox}
\end{table*}

\begin{table}
\centering
\caption{Quantitative comparison of various deraining methods on Rain200L and Rain200H datasets \textbf{Bold} is the best, \textcolor{red}{red} is the second.}
\begin{adjustbox}{width=\textwidth}
\begin{tabular}{c c c c c c c c c c}
\toprule
\multirow{2}{*}{Dataset} & \multicolumn{1}{c}{Methods} & \multicolumn{1}{c}{Input} & \multicolumn{1}{c}{SPDNet\cite{spd}} & \multicolumn{1}{c}{DualGCN\cite{dualgcn}}  & \multicolumn{1}{c}{IDT\cite{in-method3}}   & \multicolumn{1}{c}{Restormer\cite{vit2}}& \multicolumn{1}{c}{DRSformer\cite{DRSformer}} & \multicolumn{1}{c}{ESDNet\cite{ESDNet}} & \multicolumn{1}{c}{Our} \\

 & \text{Venue\&Year}  & - & \textit{(ICCV’2021)} & \textit{(AAAI’2021)} & \textit{(TPAMI’2022)} &\textit{(CVPR’2022)} & \textit{(CVPR’2023)} & \textit{(IJCAI’2024)} & - \\
\midrule
\multirow{2}{*}{200L} &  PSNR $\uparrow$        &26.70            & 40.59  & 40.89 & 41.04     & 40.80 &\textbf{41.29} &  39.85 &\textcolor{red}{41.09} \\ 
\multirow{2}{*}{}& SSIM $\uparrow$          &0.8440           & 0.9876   & \textbf{0.9892}  & 0.9885     & 0.9884 &\textcolor{red}{0.9890}  &  0.9869 &0.9886 \\ 
\midrule
\multirow{2}{*}{200H} & PSNR $\uparrow$       &13.08             & 31.30   & 31.17 & 32.03     & 32.12 &\textcolor{red}{32.19} &  30.01 &\textbf{32.62}\\ 
\multirow{2}{*}{} &SSIM  $\uparrow$         &0.3747           & 0.9212   & 0.9139 &  0.9333     & \textcolor{red}{0.9350} &0.9329 &  0.9132 & \textbf{0.9370}\\
\bottomrule
\end{tabular}
\label{table2}
\end{adjustbox}
\end{table}

\begin{figure*}[h]
    \centering
    \includegraphics[width=1\textwidth]{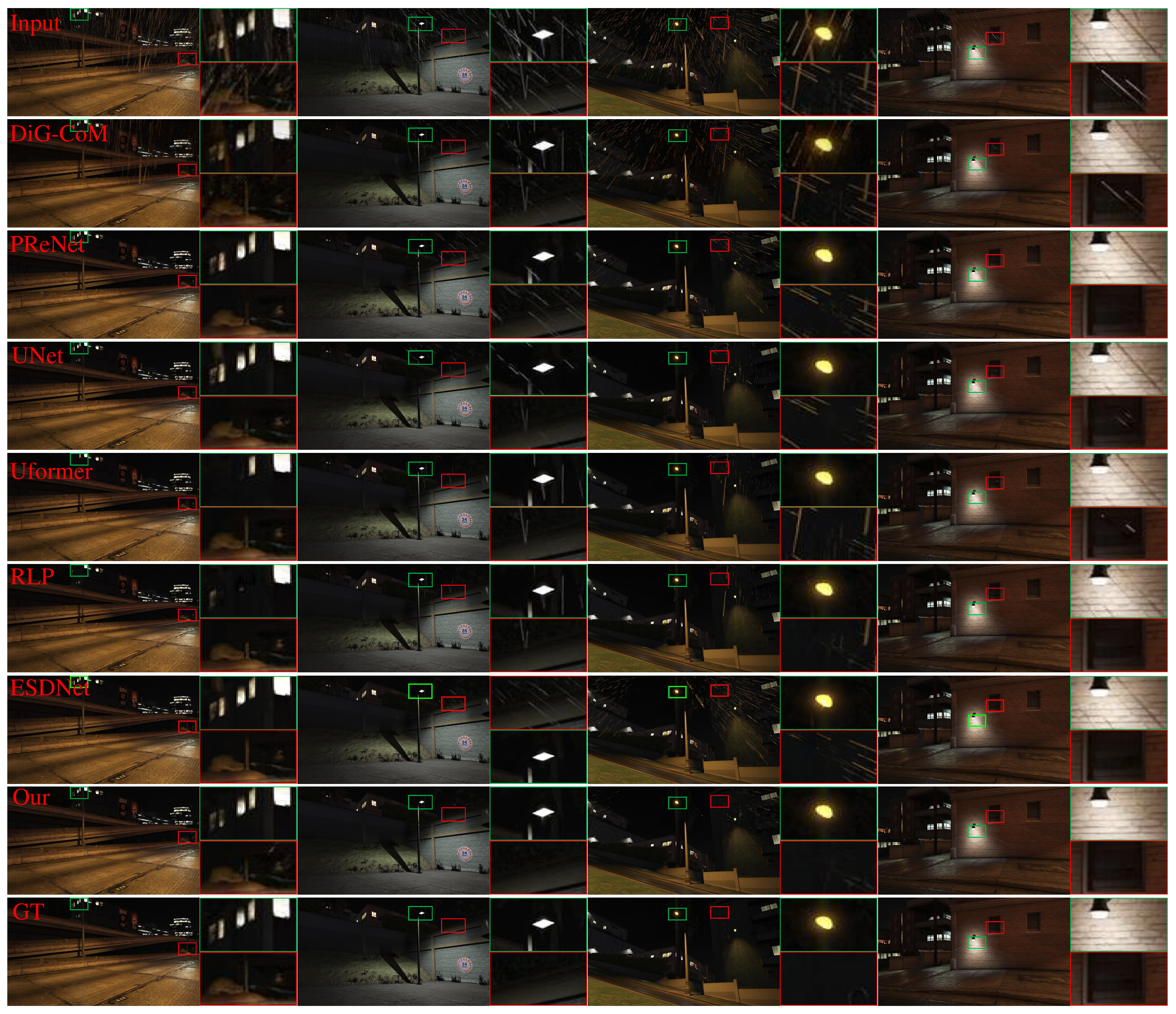}
    \caption{ Visual comparison of different deraining methods on the GTAV-Nightrain dataset, each row from top to bottom, Input, DiG-CoM, PReNet, UNet, Uformer, RLP, ESDNet, Our, GT(magnification can provide a clearer view). }
    \label{fig4}
\end{figure*}

\subsection{Experimental Results}
\begin{figure}[h]
    \centering
    \includegraphics[width=1\linewidth]{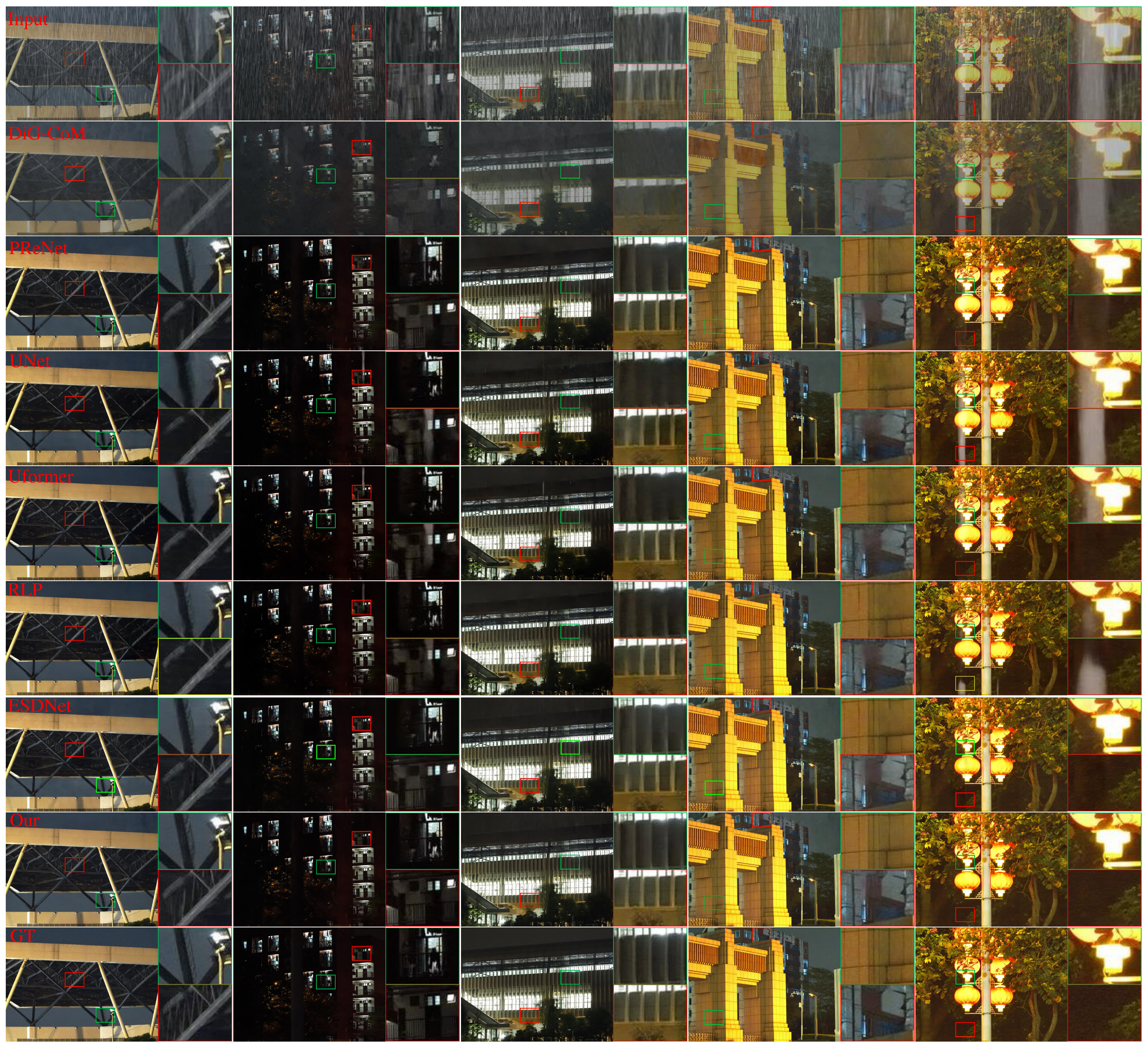}
     \caption{Visual comparison of different deraining methods on our NSR dataset, each row from top to bottom, Input, DiG-CoM, PReNet, UNet, Uformer, RLP, ESDNet, Our, GT(magnification can provide a clearer view). }
    \label{fig3}
\end{figure}

\noindent\textbf{Qualitative Results on Daytime Synthetic Data}\\
The generalization ability of the proposed method is evaluated under daytime conditions by training the model on Rain200L and Rain200H datasets. In order to reflect the diversity of the selected comparison methods, we selected other daytime rain removal SOTA methods for comparison, and the quantitative evaluation results are shown in Table \ref{table2}, our model also achieves effective rain removal during the day. On the Rain200L dataset, the denoising effect of the proposed method is better than most other methods in the daytime scene, and the PSNR/SSIM reaches 41.09/0.9886, in which the PSNR is second only to DRSformer. On the Rain200H dataset, the proposed method achieves the best performance, ranking first with PSNR/SSIM of 32.62/0.9370. 
To further confirm the effectiveness of our method, Fig.\ref{fig5} shows the visual comparison of different deraining methods on the Rain200H dataset. The rain line in Rain200L is sparse and has low difficulty, and each method performs well, which is not conducive to the observation difference. Therefore, only visual effects of different methods on Rain200H dataset are attached here. The subjective and objective experimental results show that the proposed method also has strong generalization ability in daytime scenes.
\noindent\textbf{Qualitative Results on Nighttime Synthetic Data}\\
We initially tested our method on nighttime rain-containing datasets GTA-Nightrain and NSR. Table\ref{table1} presents a quantitative evaluation of various rain removal methods on these datasets. The results indicate that on the GTA-Nightrain dataset, our method significantly outperforms other rain removal methods, achieving a PSNR that is 2.49dB higher than that of RLP. Similarly, on the NSR dataset, our proposed method surpasses all compared methods, achieving the highest PSNR/SSIM values. 
To further validate our method's performance, Fig.\ref{fig4} and \ref{fig3} provide a visual quality comparison of different rain removal methods on the GTA-Nightrain and NSR datasets. It is evident that some methods, such as PReNet and DiG-CoM, only partially remove the rain streaks. While UNet, Uformer, and RLP show better de-raining effects on the GTA-Nightrain dataset, residual rain streaks still remain. On the NSR dataset, all methods result in some blur and distortion in the recovered images. Our position perception module effectively captures the spatial locations of rain streaks, enabling more effective rain removal and better retention of original image details for higher-quality results. These subjective and objective experimental results underscore the superiority of our method and confirm its effectiveness.
\begin{figure}[t]
    \centering
    \includegraphics[width=1\linewidth]{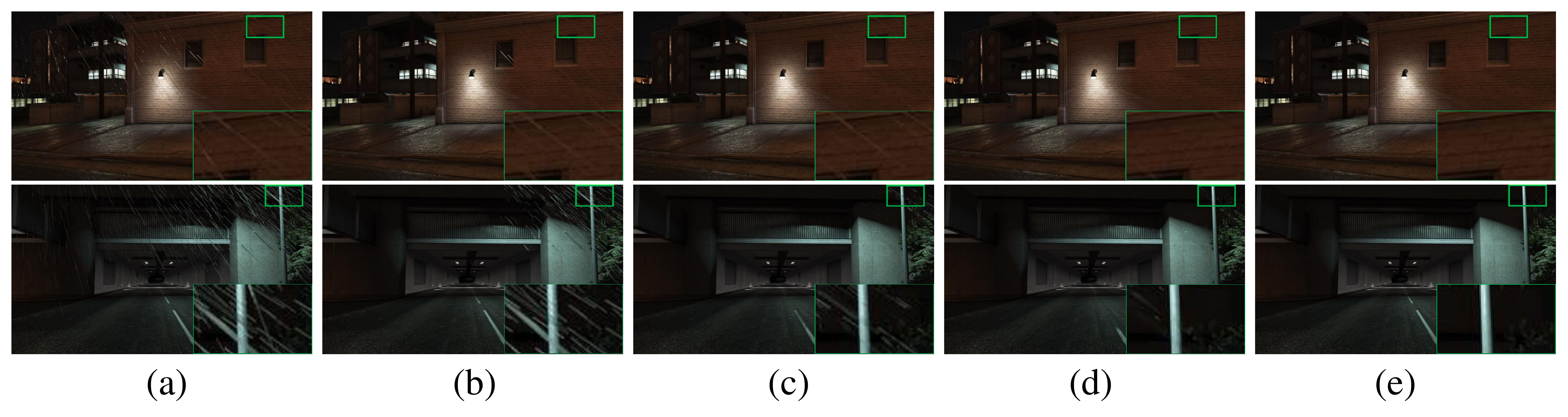}
    \vspace{-1em}
    \caption{(a) is the input rain image, (b) is the result map after ablating the PPM module, (c) is the result map after ablating the ECA, (d) is the result map of our proposed method, and (e) is the corresponding clean image.}
    \label{fig6}
    \vspace{-1em}
\end{figure}

\subsection{Ablation Study}
In this section, to validate the effectiveness of our proposed design, we conduct an ablation study on the components of our method to analyze their impact on the performance of nighttime rain removal. The configurations tested in the ablation study include:

\noindent(1) \textbf{W/O PPM}: This configuration removes the entire position Perception module(PPM) from the model to verify the effectiveness of the PPM module across the model.\\
(2) \textbf{W/O ECA}: This configuration only removes the Effective Channel Attention (ECA) module in the PPM to verify the effectiveness of the ECA module, while side-checking the effectiveness of the SPC module.\\
For these models above, we employ the same training data and settings as those used for the proposed method, utilizing the synthetic dataset GTAV-Nightrain for both training and testing. Table\ref{table3} presents the quantitative evaluation results for these configurations. The results demonstrate that each proposed module contributes to improved model performance, affirming the overall effectiveness of our design. \\
(1) The results for W/O PPM indicate that obtaining three-dimensional spatial location information on rain patterns is crucial for effective rain removal. The inclusion of the PPM module enhances the PSNR/SSIM from 37.29/0.9552 to 38.68/0.9661.\\
(2) The results for W/O ECA demonstrate that after employing spatial location encoding to capture the spatial locations of rain patterns more accurately, adding the ECA module helps the model to better focus on relevant extracted information, reduce the influence of redundant data, and boost the PSNR/SSIM from 38.44/0.9647 to 38.68/0.9661.

The visual comparison plot of the ablation models is shown in Fig.\ref{fig6}, where the green box highlights that our proposed full model achieves superior results, effectively removing rain streaks while preserving good detail in nighttime scenes. This visual comparison further verifies the effectiveness of our modules.
\begin{table}
\centering
\caption{Ablation studies using different modules on the GTAV-Nightrain dataset. \textbf{Bold} is the best, \textcolor{green}{Green} represents an increase.}
\begin{adjustbox}{width=0.5\linewidth}
\begin{tabular}{c c c c c}
\toprule
\text{Metric} & \text{PSNR} $\uparrow$ & \( \triangle \) & \text{SSIM} $\uparrow$ & \( \triangle \)  \\
\midrule
W/O PPM        &37.29            &-  & 0.9552 & -    \\ 
W/O ECA         &38.44           & \textcolor{green}{+1.15}  & 0.9647  &\textcolor{green}{+0.0095}      \\ 
Our          &\textbf{38.68}           & \textcolor{green}{+1.39}   & \textbf{0.9661}  & \textcolor{green}{+0.0109}    \\ 
\bottomrule
\end{tabular}
\label{table3}
\end{adjustbox}
\end{table}
\section{Conclusion}
We propose a location-enhanced perceptual nighttime rain removal method specifically designed for nighttime image deraining.
Considering that the location information of rain streaks in the night scene is more important than that in the day, we add the location prior of rain streaks to initially obtain the location information and distribution of rain streaks, and propose a position perception module to further extract the spatial location of rain streaks according to the rain streaks prior map. Considering that the general Transformer is not conducive to processing two-dimensional or three-dimensional data, a spatial position encoding is proposed to generate spatial position embedding and help the model to better capture and utilize the spatial context information of the input data according to the density of the distribution of rain streaks in the image. Through efficient channel attention, the model's ability to identify and adjust the importance of each channel of the feature map is further enhanced, so that the model focuses on the rain grain information and reduces the influence of redundant information. It helps the model to better preserve the fine details of the reconstructed image while accurately removing the rain. Experimental results show that our method outperforms the state of the art methods, while also generalizing well in daytime scenarios.

The proposed method is designed for nighttime deraining tasks, but it has limitations in model efficiency and requires further improvement in generalization capability for daytime scenarios. Future work will focus on optimizing these two aspects and extending the task to raindrop removal.

\section*{Acknowledgement}
This research was supported by the National Natural Science Foundation of China (No. 62201149), the Natural Science Foundation of Guangdong Province (No. 2024A1515011880), the Basic and Applied Basic Research of Guangdong Province (No. 2023A1515140077), and the Research Fund of Guangdong-HongKong-Macao Joint Laboratory for Intelligent Micro-Nano Optoelectronic Technology (No. 2020B1212030010).

 \bibliographystyle{elsarticle-num} 
 \bibliography{doc/bib/reference}







\end{document}